\documentclass{article}

\usepackage{arxiv}

\usepackage{hyperref}       
\usepackage{url}            
\usepackage{booktabs}       
\usepackage{amsfonts}       
\usepackage{nicefrac}       
\usepackage{microtype}      
\usepackage{lipsum}
\usepackage{graphicx}
\usepackage{multicol}
\graphicspath{ {./images/} }

\title{Self-Replicating Neural Programs}

\author{
 Samuel Schmidgall \\
}

\begin{document}
\maketitle
\begin{abstract}

In this work, a neural network is trained to replicate the code that trains it using only its own output as input. A paradigm for evolutionary self-replication in neural programs is introduced, where program parameters are mutated, and the ability for the program to more efficiently train itself leads to greater reproductive success. This evolutionary paradigm is demonstrated to produce more efficient learning in organisms from a setting without any explicit guidance, solely based on natural selection favoring  organisms with faster reproductive maturity.

\end{abstract}


\section{Introduction \& Related Work}


\textbf{What is self-replication? What are digital organisms?} Self-replication can be described simply as the \textit{process} by which something produces a similar or identical copy of itself, and a self-replicator is anything capable of undergoing this process. Self-replication constitutes the process by which the majority of living things are brought into creation and is a critical mechanism that both single- and multi-cellular biological organisms are dependent on. However, self-replication is a process that is not limited to biological organisms. Self-replicating computer programs have been a topic of interest for studying the dynamics of Darwinian evolution and for validating hypotheses in evolutionary theory \cite{lenski2003evolutionary, wilke2001evolution, lenski1999genome}. Evolutionary self-replicating programs, or, digital organisms, have a rich history in the fields of both experimental biology and artificial life \cite{ray1992evolution, ofria2005avida, taylor1997studying, koza1994spontaneous, reggia1998self, sipper1998fifty, reggia1998cellular}, but have however remained largely unexplored within the context of artificial intelligence .


\textbf{From digital programs to digital organisms.} Perhaps the first digital organisms emerged out of Bell Labs in 1961 from the game Darwin, where two computer programs competed for existence by trying to halt the execution of each other \cite{mcilroy1972darwin}. These computer programs were written in 7090 machine code, and had a designated set of free memory termed the arena. The game ended after either a set amount of time or once only one program remained in memory. However, the game quickly came to an end only weeks after its creation, as, one of the game's creator, Robert Morris, discovered an "ultimately lethal" program that was unable to be challenged by any competing programs.

Inspired by Darwin, Core War places two competing assembly programs in a virtual computer which battle for total control \cite{dewdney1984game}. Both programs are loaded into a random location in memory and take turns executing one instruction at a time. The game ends only once the opposing process is terminated. While many strategies for attack were developed, some of the most successful were those involving self-replication, where programs would fill the game memory with copies of itself, which in turn would produce copies of themselves. Around the same time as Core War, the first self-replicating computer programs were beginning to emerge in 'the wild,' propagating themselves across unprotected networks and embedding their self-replicating instructions into file systems -- the first digital worms and viruses \cite{cohen1986computer}. As self-replication in computer programs became a more realistic concern, so did the possibility of evolution within these programs. However, while computer viruses, worms, and Core War programs are all able to self-replicate, they are incapable of evolution.



Tierra was the first work to demonstrate that computer programs were capable of meaningful and complex evolution \cite{ray1992evolution}. The primary innovation was in designing a mutationally robust programming language, Tierran, which gave careful consideration to the way in which the representation in a programming language affects its evolvability. Tierra demonstrated the possibility for the emergence of thousands of unique programs from a single ancestral organism. Parasites evolved, which were capable of exploiting the replication code of other organisms. These parasites were eventually exploited by hyper-parasites, which used the parasite's exploitation to replicate themselves in cycle of ever-evolving digital ecology. Avida extends work in Tierra under a set of different dynamics \cite{lenski2003evolutionary}. Most notably, each program contains its own region of memory that is inaccessible to other programs, and executes its code with a separate virtual CPU. The second major difference is that organism CPU execution speed is determined by its metabolic success, measured in its ability to perform logical operations.

What Tierra and Avida so eloquently appreciate is the idea that evolution emerges as a property of reproducing organisms under the pressure of natural selection. That is to say, there is not an all-encompassing algorithm being applied to biological organisms, rather evolution emerges the organisms themselves. Current approaches in evolutionary computation do not reflect the \textit{emergence} of evolution, and rather embody the idea that evolution is applied to a population algorithmically. Self-reproduction is vital for the evolution of digital organisms, as without it, selection must be hand-designed and performed by the simulator; hence, evolutionary algorithms and evolutionary self-replicators are distinguished by a matter of \textit{natural selection} versus \textit{artificial selection} \cite{ray1992evolution, eiben2015evolutionary}.

\textbf{Self-replication and neural networks.} A recent approach toward utilizing natural selection demonstrates the emergence of intelligence on game and robotic learning benchmarks without any notion of reward based purely on natural selection \cite{schmidgall2021evolutionary}. In this work, benchmark environments are re-defined solely in terms of life and death, and organisms have a probability of replication into adjacent cells at each environment timestep. If the adjacent cells are full, then replication does not occur, and if any cells are open, in the event of a neighboring organism death, a mutated copy of the organism replicates into the adjacent location. In this way, organisms which survive longer than their neighboring organisms are more likely to produce offspring. Since the survival in each environment is defined around \textit{interesting behavior}, an organisms surviving longer inherently pertains to better environment performance, and in turn, many of the organism populations attain performance competitive to reward-based optimizers without any notion of reward. In this way, local interactions between replicators produces the emergence of evolution in traits pertaining to survival in their respective environment. However, the explicit process of replication is performed through memory copying. In this way, the digital organisms were not replicating of their own will, or from their own internal capabilities, but merely based on a simulator-defined probability distribution.

Making the process of replication more explicit, Neural Network Quines \cite{chang2018neural} are neural networks trained to output their own weights, with the training mechanism focusing solely on reducing this predictive discrepancy. The outcome of an optimally trained neural network quine, is a network capable of reproducing its own weights. The reproduced weights could then be applied to another neural network of the same topology, and that network would ideally be capable of reproducing itself. Since it is not entirely possible for a network to output its own weights in their entirety, as every output neuron producing a weight requires at least one additional weight in the network, neural network quines utilize an indirect encoding to map network outputs to network weights. While these networks could under some definitions be considered self-replicators, neural network quines are dependent on external mechanisms for replication and training that the quine itself does not produce; the total machinery necessary for replication does not exist within the replicator. In this way, it is difficult to argue that neural network quines are truly self-replicating. Additionally, by only training to produce its own weights, the network is incapable producing any form of evolution on its own.

Toward improving upon this, the self-replicating neural networks presented in this work, by themselves, contain all of the information necessary to construct a copy of the \textit{mechanism} by which they are derived. The neural program has no dependencies other than itself while undergoing replication, using its own generated output as input. It is shown that these self-replicating neural programs are capable of undergoing evolution in the context of natural selection, where the training time for a given task is shown to decrease over time while replication occurs in an unstructured manner.

\begin{figure*}%
    \centering
    \includegraphics[width=16.5cm]{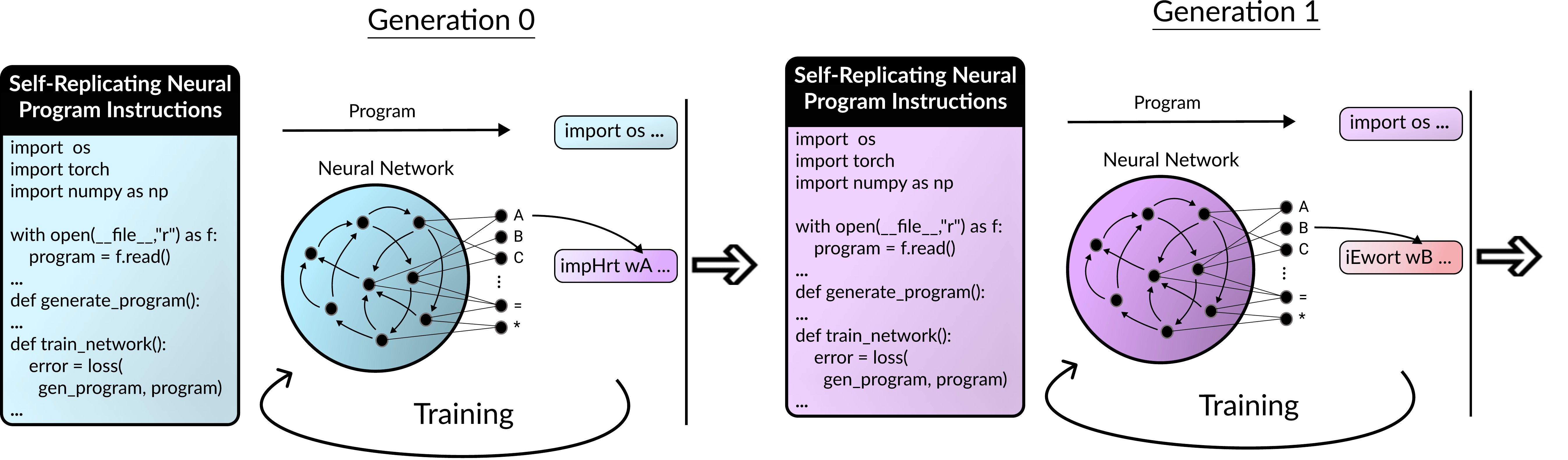}
    \caption{Depiction of a basic self-replicating neural program. Generation 0 begins with code responsible for constructing and training a neural network to generate its own code, and once training is finished, and the replicator is capable of producing a replica of itself, the program writes and executes a copy of the neural program on to Generation 1, ad infinitum.}%
    \label{fig:basic_program}%
\end{figure*}

\section{Experiments}

\subsection{Self-Replicating Neural Program}
This process begins by the program reading and storing a copy of its own instructions into memory. The goal is to produce a network capable of generating these instructions. The generation of programs serves as a useful representation for replication, as a significant amount of information can be stored in a relatively small amount of program text. Since there are a finite number of possible characters used to represent program text, $N_{p}$, it is useful to represent the network output vector with $N_{p}$ neurons representing each possible character in the program. The input vector is not as trivial to represent, as the goal is for the network to produce a copy of itself without referring to the program string during the generation process. Toward this end, the $N_{p}$ dimensional network output values are used as input for the next character generation, with the first input in the generation process represented as a one-valued vector. Once an entire sequence has been generated, it is compared with the set of instructions stored in memory, and the network parameters are adjusted to more closely match these program instructions. When the network successfully produces a copy of its own instructions, this copy is written to a new file and then executed, which is all accomplished within the original instruction set itself. This newly executed program then goes on to train its own self-replicating neural program, carrying on the process for generations ad infinitum. 

While the process of training is agnostic to any particular learning algorithm, backpropagation through time with stochastic gradient descent applied to the Adam optimizer \cite{kingma2017adam} with a learning rate of $0.001$ is used to minimize the loss in this instance. The neural network used to generate the program is represented by a 2-hidden layer LSTM with 64 cells in each layer. The network output vector is the dimensionality of the number of possible characters plus a set of auxiliary neurons. The auxiliary neurons allow for the network to learn signals can be used in the input layer at the next character generation timestep. The generated character during training is sampled categorically using a softmax on the character neuron output vector for the category probabilities. During reproductive evaluation, the generated character is taken from the character neuron with the maximum value. The input information at generation time $t$ is the output of the network at time $t-1$ along with a set of $N=8$ random-noise neurons sampled from a normal distribution. At the first generation timestep $t = 0$, the output vector portion of the input is represented by a one-vector.

However, despite the capability for neural self-replication, the fact remains that computers are capable of trivial replication in the form of memory copying, and the utility of a replicating neural program is brought into question. Presented below are several motivations for the study of self-replicating neural programs, some of which are further explored in this paper.



\textbf{Evolutionary Self-Replicating Neural Programs} \vspace{-4px} 

Self-replication and reproduction in the context of natural selection has led to many of the incredible accomplishments of life that exist today. Evolution in the form of natural selection has been shown to produce intelligence not only in biological organisms, but also in digital organisms \cite{lenski2003evolutionary, ray1992evolution, schmidgall2021evolutionary}. A paradigm for evolutionary self-replication in neural programs is presented in this work to demonstrate this possibility.

\textbf{Metacognitive Self-Awareness and Self-Improvement} \vspace{-4px}  

The process of learning a representation of oneself opens up many questions about self-awareness. Is it possible for a program to have an awareness about itself, potentially shaping modifications in itself or in its replicated copies, and observing the outcomes to determine how to improve its program? This type of improvement differs from evolutionary self-replication, which depends on random mutation and natural selection for unguided improvement. Self-aware improvement, or self-improvement, offers itself as an attractive potential for the future of autonomous meta-learning algorithms.

\textbf{Programmatic Self-Repair} \vspace{-4px}

In practical situations, it is not difficult to imagine self-repair as being a useful mechanism for a self-sufficient program. Self-repair is fundamental in biological organisms, as even small damages that remain unrepaired can prove fatal. In the computational setting, instructions can be damaged by many factors, including alpha particles, cosmic rays, thermal neutrons, and hardware defects. Additionally, competing programs could evolve the ability to damage the functionality of other programs if it proves evolutionarily advantageous \cite{ray1992evolution}. Neural programs provide a natural solution to self-repair, as their instructions learn to generate a representation of itself, and hence are trivially capable of regeneration.

\textbf{Autonomous Programs} \vspace{-4px}  

Self-replicating neural programs have the potential to initialize intelligence that controls the modification, maintenance, and design its own code, as well as the creation of new code fully autonomously -- intelligence that exists and develops itself entirely outside of the control of humans. By limiting the number of programs that can exist, and introducing a notion of natural selection, it is possible for these self-replicating neural programs to be driven toward a goal within their evolutionary progression that leads to novel program designs.

\textbf{The Possibility of Neural Malware} \vspace{-4px} 

By condensing a program into the weights of a neural network and training it tabla-rasa starting with random weights every generation cycle, it is likely that each produced network will have a unique set of learned parameters. Observing the structure and parameters of a neural network does not provide enough information to understand it's functionality \cite{cohen1986computer}; as a consequence, it is possible for the nature of neural programs in their condensed form to remain elusive as they propagate through digital space. This becomes increasingly plausible when considering that the most common form of malware detection is signature-based, requiring an exact digital "footprint" within the program's instructions, which in the case of neural programs, lies resident inside the uninterpretable network parameters. Early theoretical analysis claims that the "detection of a virus is shown to be undecidable both by a-priori and runtime analysis, and without detection, cure is likely to be difficult or impossible \cite{cohen1986computer}." It is also unlikely that all arbitrary neural structures would be flagged, especially as the application cases for neural networks continues to rise, leaving the possibility of camouflaged self-replicating neural programs. These programs could be capable of evolution across generations and intelligent decision making during their lifetime, unlike any malware existing today. It is important to consider and study such possibilities before the potential becomes a reality.


\subsection{Evolutionary Self-Replicating Neural Programs}

The ability for a self-replicating organism to undergo change across generations is critical for enabling the occurrence of evolution; otherwise, replicated organisms would remain identical. In the previous examples of self-replicating neural programs, replication produces an exact copy of the organism's instructions, preventing the potential for evolutionary processes to occur. Here it is shown that mutation and natural selection can be introduced into the replication process, ultimately leading to faster replication times across generations.

\begin{figure*}%
    \centering
    \includegraphics[width=\linewidth]{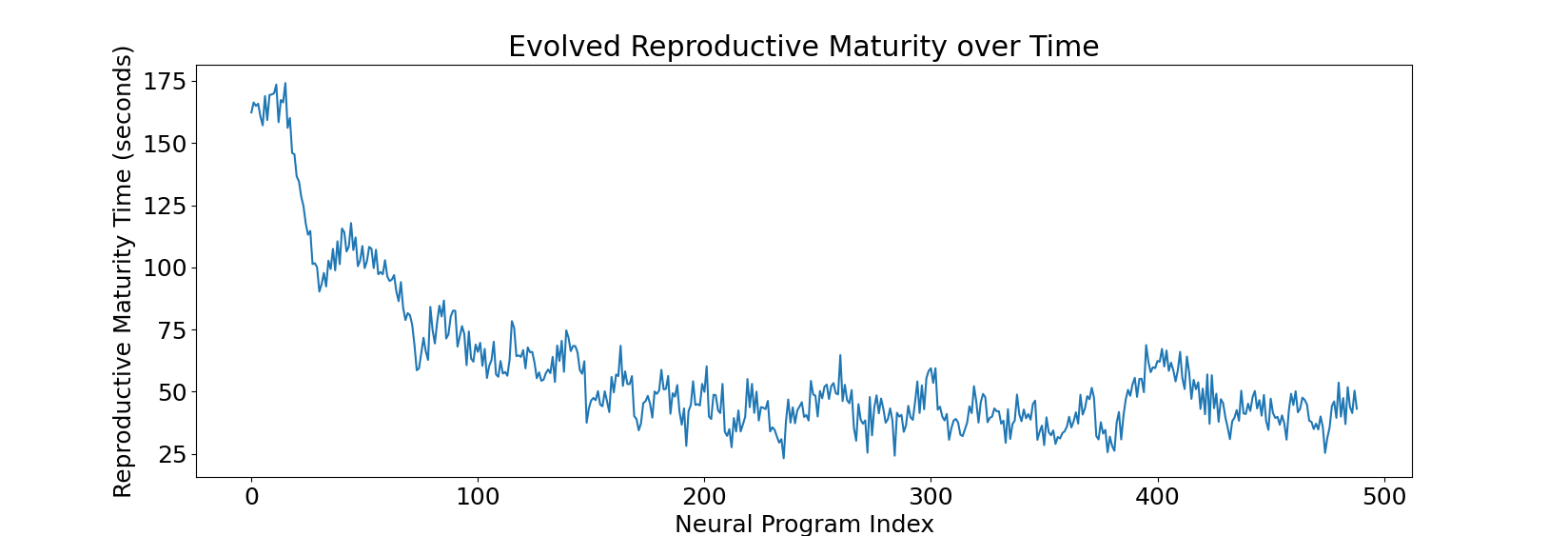}
    \caption{Depiction of the average time (seconds) taken to train each neural program in the population (neural program index). Neural programs are indexed by their respective completion time.}
    \label{fg:evo_maturity}%
\end{figure*}

To begin the evolutionary process, a single self-replicating program is initialized, called the ancestor program \cite{ray1992evolution}. In the program demonstrated in Section 2.1, each neural program replicated a single copy of itself before termination. However, for evolution to occur without centralized algorithms, successfully reproducing programs need more than a single offspring with some form of variation. To meet these qualifications, each self-replicating program is modified to produce two offspring, each of which containing a mutated copy of the parent program. Mutations occur on the network learning rate, which is sampled from a zero-centered Gaussian distribution, and on the number of hidden neurons, input noise neurons, and peripheral output neurons, which was sampled from a categorical distribution between $-5$ and $5$.

While it would be trivially simple to programmatically select the fittest candidates according to some defined metric or fitness function, it is more interesting to observe evolution occurring without any form of intervention. However, for evolution to occur, there must be \textit{some} form of selection, as programs would otherwise self-replicate without any means of improvement, generating an exponentially growing population in the process. On computers however, population growth occurring without bounds is not entirely realistic, as digital computers have a finite number of spatial resources, where once these resources begin to run out, other programs in storage are overwritten. To emulate the natural process of population growth in digital computers, while simultaneously enabling evolution, a fixed amount of resources are reserved for self-replicating organisms, and once the space fills up, programs that replicate themselves overwrite others in storage. In this way, programs with the fastest replication process are able to, with enough time, out-compete those with slower replication, naturally evolving a population of increasingly sophisticated replicators without any form of explicit selection.

To do this, a high level program is defined which keeps track of current neural program process IDs (PIDs) active in the kernel. This program periodically checks whether new processes are present (replicated offspring) or are absent (parent program finishes replication and ends execution). If a new process is presented, and the maximum number of processes are already in session, the oldest process in the queue is killed, and replication does not occur. In this way, it acts as if the new process is overwriting the older one, introducing the notion of finite resources to enable natural selection pressuring faster replication. It is crucial to understand that these programs are not operating centerally, where each program is writing and executing their own children through internal commands. In this way, the neural programs are capable of operating entirely independent of centralized organization -- they are in charge of their own agency.

For the experiment displayed in Figure \ref{fg:evo_maturity}, the maximum number of simultaneous programs was limited to 16. The initial set of programs took on average 175 seconds before being able to produce a copy of itself. Across the lifespan of 500 neural programs in competition, this average was reduced to 41 seconds -- only requiring $23$\% of the initial time. The initial learning rate in the ancestor program was $0.001$, which increased and stabilized around the value $0.015$ for the more rapid replicators. Additionally, the number of hidden neuron increased from 16 to an average of 31, whereas the noise and auxiliary output dimensions did not seem to increase significantly. Overall, neural programs are capable of evolving meaningfully across generations without any form of explicit fitness functions and selection techniques applied to the population. This demonstrates that all of the ingredients necessary for the emergence of meaningful evolution can be derived from the properties of a single self-replicating organism.



\section{Future work and discussion}

In this work, a paradigm is presented where a neural network is trained to output the code that trains it using only its input as output. In this way, the self-replicating neural programs by themselves are capable of producing a copy of the mechanism by which they are derived. The replication process of these neural programs are modified to produce a mutated copy of the program parameters, which, in competition with other neural programs, is shown to produce more efficient learning in a setting without objectives or guidance -- natural selection. The act of self-replication was argued to have several potential utilities, including metacognitive self-awareness and self-improvement, programmatic self-repair, autonomy, and as was demonstrated experimentally, the ability for evolutionary self-replication. There are also potential dangers that follow the existence of neural programs, such as neural malware, which is argued to pose a threat to modern footprint-based malware detection systems.

While self-replication under the pressure of replication speed was shown as the focus of evolution in this work, it has also been demonstrated that the same process may be used to solve Atari and robotic learning tasks in neural networks \cite{schmidgall2021evolutionary}. In that work, however, the process of replication was abstracted, whereas this work makes replication explicit. Perhaps future work will both produce the capability for replication in the form of neural programs, and task solution in the form of evolutionary replication. Additionally, the evolution of neural programs presented in this work has a limited domain of possible programs that can be produced, as only a fixed number of targeted hyperparameters were mutated. While these programs were shown to evolve meaningfully toward faster training times, the fundamental nature of their behavior remains the same across generations, as the general structure of the training remains identical. It would be interesting if these programs could change the fundamental structure of their behavior, or even evolve functionality across multiple programming languages.



\bibliographystyle{unsrt}
\bibliography{references}

\end{document}